\documentclass[sigconf]{acmart}

\AtBeginDocument{%
  }

\setcopyright{acmcopyright}
\copyrightyear{2018}
\acmYear{2018}
\acmDOI{XXXXXXX.XXXXXXX}

\acmConference[Conference acronym 'XX]{Make sure to enter the correct
  conference title from your rights confirmation emai}{June 03--05,
  2018}{Woodstock, NY}
\acmPrice{15.00}
\acmISBN{978-1-4503-XXXX-X/18/06}

\usepackage{romannum}
\usepackage{amsmath}
\usepackage{amsfonts}
\usepackage{bm}
\usepackage{multirow}
\usepackage{algorithm}
\usepackage{algorithmic}
\usepackage{caption}
\usepackage{balance}
\usepackage{subcaption}
\usepackage{colortbl}
\usepackage{graphicx}
\usepackage{float}
\usepackage{wrapfig}

\usepackage{tikz}

\usepackage{url}            
\usepackage{booktabs}       
\usepackage{amsfonts}       
\usepackage{nicefrac}       
\usepackage{microtype}      

\usepackage{xcolor}         

\copyrightyear{2023}
\acmYear{2023}
\setcopyright{acmlicensed}\acmConference[CIKM '23]{Proceedings of the 32nd ACM International Conference on Information and Knowledge Management}{October 21--25, 2023}{Birmingham, United Kingdom}
\acmBooktitle{Proceedings of the 32nd ACM International Conference on Information and Knowledge Management (CIKM '23), October 21--25, 2023, Birmingham, United Kingdom}
\acmPrice{15.00}
\acmDOI{10.1145/3583780.3615267}
\acmISBN{979-8-4007-0124-5/23/10}

\begin{document}

\title{T-SaS: Toward Shift-aware Dynamic Adaptation for Streaming Data}


\author{Weijieying Ren}
\email{wjr5337@psu.edu}

\affiliation{%
  \institution{The Pennsylvania State University}
  \city{State College}
  \state{Pennsylvania}
  \country{USA}
}

\author{Tianxiang Zhao \footnotemark[1]}
\email{tkz5084@psu.edu}
\affiliation{%
  \institution{The Pennsylvania State University}
  \city{State College}
  \state{Pennsylvania}
  \country{USA}
}

\author{Wei Qin}
\email{qinwei.hfut@gmail.com}
\affiliation{%
  \institution{Hefei University of Technology}
  \city{Hefei}
  \state{Anhui}
  \country{China}
}

\author{Kunpeng Liu}
\email{kunpeng@pdx.edu}
\affiliation{%
  \institution{Portland State University}
  \city{Portland}
  \state{Oregon}
  \country{China}
}

\renewcommand{\shortauthors}{Weijieying Ren, Tianxiang Zhao, Wei Qin,  Kunpeng Liu}



\vspace{-5cm}

\ccsdesc[500]{Computing methodologies~Neural networks}
\ccsdesc[500]{Computing methodologies~Learning latent representations}
\ccsdesc[500]{Computing methodologies~Classification and regression trees}
\keywords{data stream, model generalization, distribution shift}
\vspace{-5cm}



\begin{abstract}
In many real-world scenarios, distribution shifts exist in the streaming data across time steps. Many complex sequential data can be effectively divided into distinct regimes that exhibit persistent dynamics. Discovering the shifted behaviors and the evolving patterns underlying the streaming data are important to understand the dynamic system. Existing methods typically train one robust model to work for the evolving data of distinct distributions or sequentially adapt the model utilizing explicitly given regime boundaries. However, there are two challenges: (1) shifts in data streams could happen drastically and abruptly without precursors. Boundaries of distribution shifts are usually unavailable, and (2) training a shared model for all domains could fail to capture varying patterns. 
This paper aims to solve the problem of sequential data modeling in the presence of sudden distribution shifts that occur without any precursors.  
Specifically, we design a Bayesian framework, dubbed as T-SaS, with a discrete distribution-modeling variable to capture abrupt shifts of data. Then, we design a model that enable adaptation with dynamic network selection conditioned on that discrete variable.
The proposed method learns specific model parameters for each distribution by learning which neurons should be activated in the full network. A dynamic masking strategy is adopted here to support inter-distribution transfer through the overlapping of a set of sparse networks.
Extensive experiments show that our proposed method is superior in both accurately detecting shift boundaries to get segments of varying distributions and effectively adapting to downstream forecast or classification tasks. 
\end{abstract}

\maketitle
\footnotetext[1]{Corresponding author}
\section{Introduction}
Deploying machine learning models \cite{ren2017robust,xu2018enhancing,zhao2020balancing,zhao2023faithful} in real-world systems often presents the challenge of distribution shift. This occurs when the statistical characteristics of newly incoming data differ from those observed by the model in a dynamically changing environment \cite{wang2018deep,de2021continual,ren2018tracking,zhao2022synthetic,ren2022mitigating}.
Generally, the shift can happen without any precursor, be unknown to users, cause dramatic personal injury for systems like self-driving \cite{liu2020learning}, robotics \cite{zhang2021adversarial,zhao2023skill}, sleep tracking \cite{wilson2020survey} and irreparable economic damage on financial trading algorithms \cite{wilson2020survey,ren2021cross,wang2022explanation,zhao2022consistency}. 
At any moment, the model is expected to 1) provide early warnings when the data distribution changes and 2) make accurate predictions by adapting to new data. 
Several approaches have been proposed to develop adaptive models for sequential data in dynamic environments, e.g., transfer learning \cite{wang2018deep,wilson2020survey,ren2022semi,zhao2022topoimb}, continual learning \cite{nguyen2017variational,shen2023graph}.
However, they are unable to deal with sudden or irregular shifts due to parameter-sharing strategies, which limits their model capacity. Besides, these algorithms mainly assume that the data stream is explicitly divided into different regimes \cite{wang2018deep} (\textbf{tasks} or \textbf{domains}) according to given change points. In the real world, changes often occur without precursors and explicit temporal segments are unavailable.

Addressing these challenges, we propose an incremental model selection approach in the form of a dynamic network to handle sequentially shifting data. 
In this paper, we introduce T-SaS, a novel Bayesian-based approach that combines dynamic network selection with a shift points estimation scheme for sequentially evolving data \cite{ren2021fair}.
This approach enables the modeling of distinct distributions by evolving the network structure \cite{zhao2023towards} and facilitates positive knowledge transfer across complex dynamic regimes. 
By back-propagating through the change point estimation, our proposed method learns a predictive model that can both capture the underlying distribution changes and quickly adapt to it. 
We leverage the principle that each specific data distribution should correspond to a sparse subset of the computation graph  \cite{zhao2018zero} of the full neural network, and adaptation to shifts can be modeled as a new network selection strategy. Different related tasks share different subsets of model parameters and follow corresponding dynamic routes along the obtained network structures. 
To incorporate automatic distribution shift detection,
we introduce a discrete change variable to estimate the compatibility between the current and previous data distributions. 
The main contribution of this paper are as follows:
\begin{itemize}
\item We propose a novel Bayesian-based method that can detect abrupt distribution shifts automatically and enable knowledge adaption on the evolving changed data.
\item We utilize change point detection to estimate abrupt distribution changes and propose a dynamic masking strategy to learn specific model parameters for
each distribution.
\item We evaluate our proposed method on both real-world and synthetic datasets
in terms of forecasting and classification problems.
We demonstrate the proposed method is capable of detecting distribution changes and adapting to evolving data through experiments on various datasets.
\end{itemize}

\section{Methodology}
Let $\{\mathcal{D}_t\}_{i=1}^{T}$ be a sequentially arriving dataset for training with each $\mathcal{D}_t$ containing 
labeled pairs $(\mathbf{X}_t,\mathbf{Y}_t)$.
At each time, the continual learner aims to optimally adapt the model parameters into $\boldsymbol{\omega}_t$ for newly incoming data $\mathcal{D}_t$ by borrowing statistical strength from previous data.
The goal of our problem setting is to train a forecasting or classification model that trains on $\{\mathcal{D}_t\}_{i=1}^{T}$ and can generalized well to $M$ arriving data $\{\mathcal{D}_t\}_{i=T+1}^{M}$. For the classification model, the input-output pairs are both given in each $\mathcal{D}_t$. For the classification model, each $\mathcal{D}_t$ is represented as $\mathcal{D}_t = \mathbf{X}_t$.

Here, we propose to infuse historical knowledge via posterior of parameters in a Bayesian online learning framework \cite{nguyen2017variational}. Concretely, the parameter posterior of $t$-th dataset is modeled as:
\begin{equation}
p(\boldsymbol{\bm{\omega}_t}|\mathcal{D}_{1:t}) \propto \int p(\mathcal{D}_t|\bm{\omega}_t)p(\bm{\omega}_t|\mathcal{D}_{1:t-1}),
    \label{Eq::bayesian}
\end{equation}
Two fundamental elements build up our research problem. First, $p(\mathcal{D}_t)$ can be non-stationary and is subject to distribution shift. 
Second, domain segmentations are usually unknown. 
To solve the two challenges, we design a ShiftNet that simultaneously detects distribution shift as well as adapts the continual learner to the inferred changes with an adaptive model structure. 
\subsection{Latent Shift-Oriented Model Design}
Addressing the aforementioned challenges, we design an adaptive network that accounts for shifts in the data distributions with a change point detection module and a structure adaptation module. 
For every incoming dataset $\mathcal{D}_{t}$, we need to decide which previous patterns in $\mathcal{D}_{1:t-1}$ that $\mathcal{D}_{t}$ is more compatible to. To this end, we introduce a new discrete variable $\mathbf{s}_t$ that accounts for distribution shift. 
The continual learner can be parameterized as :
\begin{equation}
    p(\bm{\omega},\bm{s}_t|\mathcal{D}_{1;t}) \propto p(\mathcal{D_{t}|\bm{\omega},s_{t}})p(\bm{\omega},s_{t}|\mathcal{D}_{1:t-1}).
    \label{Eq::discrete}
\end{equation}

\begin{figure}[t]
  \centering
\includegraphics[width=\linewidth]{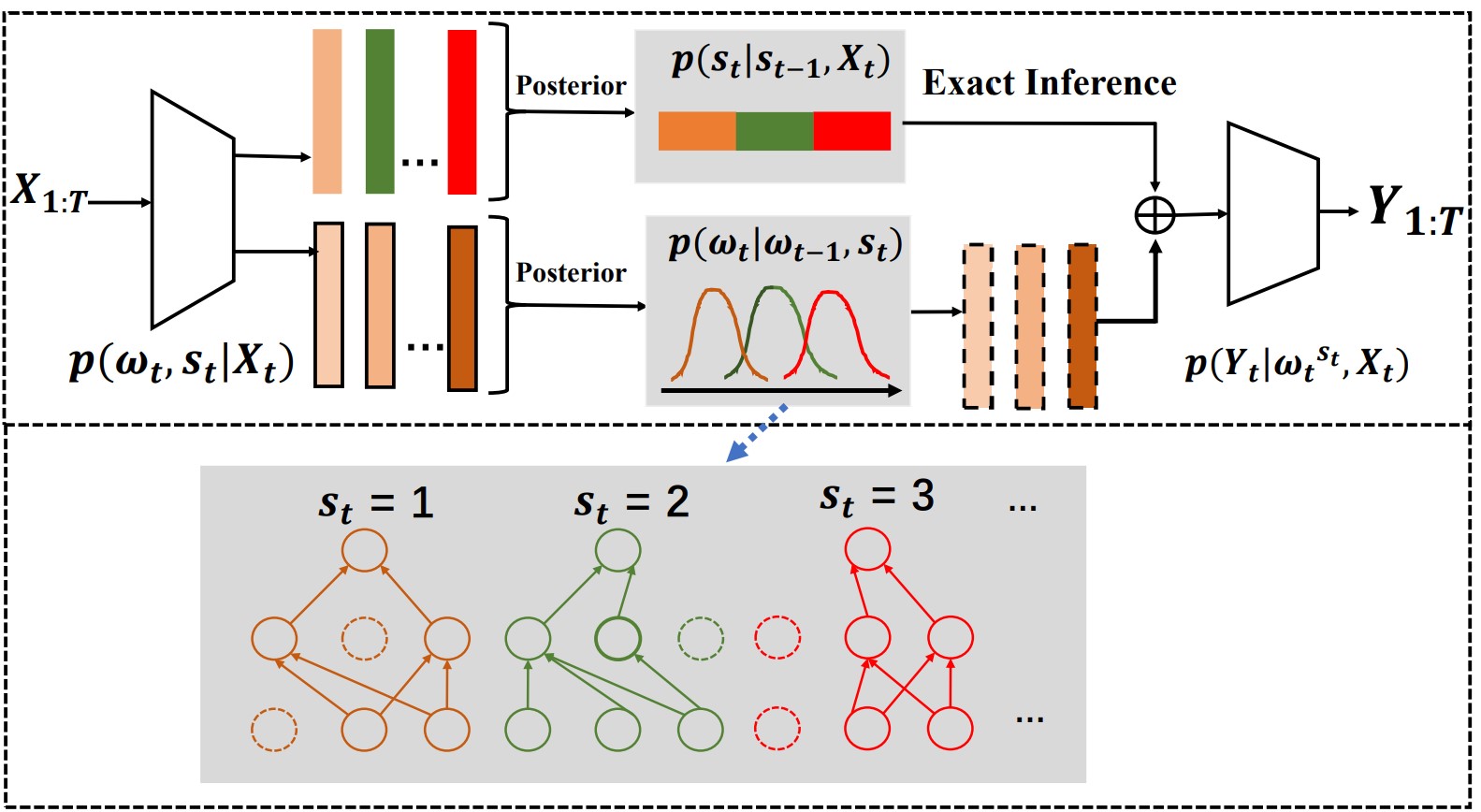}
\caption{Overview of the T-SaS method. We derive a probabilistic framework to infer the distribution shift and modulate the adaptive model simultaneously. Input $\mathbf{X}_{1:T}$ is feed into a neural network to infer two variables, i.e., the latent change point variable $\bm{s}_t$ and model posterior parameters $\bm{\omega}_t$. The change point variable is conditioned on $\bm{s}_{t-1}$ and $\mathbf{X}_{t-1}$ which encodes the time-dependent relations and can be calculated by exact inference. $\bm{\omega}_{t}$ is sampled with a dynamic masking-based strategy to determine the model prediction $p(\mathbf{Y}_{t}|\mathbf{X}_{t},\bm{\omega}_{t})$.}
\label{Fig:overview}
\vspace{-0.1cm}
\end{figure}

Note that discrete change variables $s_{t}$ influence the adaptive model parameters $\bm{\omega}$ and are indicative for sudden distribution shift. Intuitively, $s_{t}$ depends on previous change variable $s_{t-1}$ as time-varying data usually maintains a local smoothness property for time duration \cite{ansari2021deep}, but also depends on previous observation $\mathcal{D}_{t-1}$. 
This means the variable $s_{t}$ may be triggered by signals from the environment. 
To better express the flexible transition of the change point variable, we reformulate Eq.\ref{Eq::discrete} as follows:

\begin{equation}  \label{Eq::change_point}
\begin{aligned}
        &p(\bm{\omega}_{t},\bm{s}_{t}|\mathcal{D}_{1;t}) \\
         \propto & p(\mathcal[D_{t}|\bm{\omega},s_{t}])p(\bm{\omega},s_{t}|\mathcal{D}_{1:t-1})p(s_{t}|\mathcal{D}_{1:t-1},s_{t-1}).
\end{aligned}
\end{equation}
Summing over time steps, we can reformulate each term in Eq.\ref{Eq::discrete} using Markov Chain to model the joint distribution as:

\begin{equation} \label{eq::markov}
\begin{aligned}
    & p(\mathcal{D}_{1:T},\bm{\omega}_{1:T},\bm{s}_{1:T}) \\
    =& \prod_{t=2}^{T} p(\mathcal{D}_{t}|\mathcal{\omega}_t, \bm{s}_t)p(\bm{\omega}_t|\bm{\omega}_{t-1},\bm{s}_t)p(\bm{s}_t|\bm{s}_{t-1},\mathcal{D}_{t-1}).
\end{aligned}
\end{equation}
In this framework, with $\mathcal{D}_t$ being modeled as $p(\mathcal{D}_{t}|\mathcal{\bm{\omega}}_t, \bm{s}_t)$, shifts in data distribution can be encoded with change point indicator $\bm{s}_t$ for the problem of evolving data generalization. Next, we will go into details on the detection module with detected shifts.
\subsection{Change Point Detection for Data Stream}
When training with a complex dataset, the posterior distribution is intractable and needs to be inferred \cite{dong2020collapsed}. 
A common approach to learning latent variables is to maximize the evidence lower bound (ELBO) on the log marginal likelihood \cite{nguyen2017variational}. 
This is given by $\log p(\mathcal{D})\leq \mathbb{E}_{q_{\bm{\phi}}(\bm{w},\bm{s})}[\log p_{\bm{\theta}}(\mathcal{D},\bm{w},\bm{s})-\log q_{\bm{\phi}}(\bm{w},\bm{s}|\mathcal{D})]$, where $p_{\bm{\theta}}(\mathcal{D},\bm{w},\bm{s})$ shows the generative process and $q_{\bm{\phi}}(\bm{w},\bm{s}|\mathcal{D})$ is an approximation posterior. We define the prior distribution in Eq.\ref{eq::markov} for change point variable $\bm{s}$ and model parameter $\bm{w}$ as:
\begin{equation}
\begin{aligned}
    &p(\bm{w}^{\bm{s}_t}_t) \sim \mathcal{N}(\bm{w}^{\bm{s}_t}|\bm{\mu}^{\bm{s}_t},\bm{\Sigma}^{\bm{s}_t}), \\
    &p(\bm{s}_t) \sim Cat(\bm{s}_t|S(f(\bm{s}_{t-1},\mathcal{D}_{t-1}))),
\end{aligned}
\end{equation}
where $\mathcal{N}$($\cdot$) is a multivariate Gaussian distribution with mean and Var as $\bm{\mu}^{\bm{s}_t}$ and $\bm{\Sigma}^{\bm{s}_t}$. $f$ is an non-linear function (MLPs),  $\mathcal{C}at$($\cdot$) denotes a categorical distribution and S($\cdot$) is a softmax function.

To approximate the prior distribution, we propose to use variational inference to learn an approximated posterior distribution $q_{\bm{\phi}}$ over the latent variable and formulate it as: 
\begin{equation}
    q_{\bm{\phi}}(\bm{w},\bm{s}|\bm{x}) = p_{\bm{\theta}}(\bm{s}|\bm{x}) q_{\bm{\phi}}(\bm{w}|\bm{x},\bm{s}),
\end{equation}
where $p_{\bm{\omega}}(\bm{s}|\mathcal{x})$ is the exact posterior computed by forward-backward algorithm \cite{dong2020collapsed} and $q_{\bm{\phi}}(\bm{w}|\bm{x},\bm{s})$ is an approximated posterior. At this moment, we assume the distribution of  $q_{\bm{\phi}}(\bm{w}|\bm{x},\bm{s})$ is given. 
Optimizing the variational parameters $\bm{\phi}$ corresponds to minimizing the ELBO at each time step $t$:
\begin{equation} \label{Eq::likelihood}
    \begin{aligned}
    \mathcal{L}(\bm{\phi}_{t}) 
    = 
    &\mathbb{E}_
    {q_{\bm{\phi}_{t}}(\bm{w},\bm{s}|\mathbf{X})
    p_{\bm{\omega}}(\bm{s}|\bm{w},\mathbf{X})}
    [-\log p(\mathbf{X}_t|\bm{w}^{\bm{s}_t}_t)
    p_{\bm{\omega}}(\bm{s}_t|\mathbf{X}_t)] \\
    &+\log q_{\bm{\phi}_{t}}(\bm{w},\bm{s}|\mathbf{X})
    p_{\bm{\omega}}(\bm{s}_t|\mathbf{X}_t)\\
    = & \mathbb{E}_{q_{\bm{\phi}_{t}}(\bm{w},\bm{s}|\mathbf{X})}[-\log p(\mathcal{D}_t|\bm{w}^{\bm{s}_t}_t)] \\
    & + \mathbf{KL}(q_{\bm{\phi}_{t}}(\bm{w}^{\bm{s}_t}|\bm{X},\bm{s})||q_{\bm{\phi}_{t-1}}(\bm{w}^{\bm{s}_t}|\bm{X},\bm{s})),
\end{aligned}
\end{equation}
where the likelihood term $\mathbb{E}_{\mathbf{X} \thicksim {\mathcal{D}_t}}[\log p(\mathbf{X}_t|\bm{w}_t,\mathbf{X}_{t-1})]$
can be computed using the forward variable $\alpha_t(\bm{s}_t)$ by marginalizing out the change point variable $\bm{s}_t$,
\begin{equation} \label{Eq::final_eq}
\begin{aligned}
    p(\mathbf{X}_{1:T},\bm{w}_{1:T}) 
    = \sum_{\bm{s}_t}\alpha_t(\bm{s}_t)
    = \sum_{t=2}^{T}\sum_{j,k}\gamma_t(j,k)[\log B_t(k)A_t(j,k)],
\end{aligned}
\end{equation}
where $A_(j,k) = p(\bm{s}_t=j|\bm{s}_{t-1}=k,\mathbf{X}_{t-1})$ and $B_t(k) = p(\mathbf{X}_t|\bm{\omega_t})\cdot\\
p(\bm{\omega}_t|\omega_{t-1},\bm{s}_t=k)$. Here, change point variables can be calculated from exact inference.
The ELBO  can be approximated via stochastic gradient ascent given that $q_{\bm{\phi}_{t}}(\bm{w}^{\bm{s}_t}|\bm{X},\bm{s})$ is reparameterizable. 

\subsection{Structure Adaptation With Change Points}
To improve the adaption on a sequence of evolving data distribution, the continual learner is equipped with a mixture of dynamically updated model parameters $\{\bm{\omega}_t^{\bm{s}_t}\}_{\bm{s}_t=1}^K$, where $K$ is the number of categorical data numbers.
Each change point variable $\bm{s}_t$ is associated with a cluster of similar data distributions, and we define $\bm{\omega}_t^{\bm{s}_t}$ as the specific model parameters. 
Notably, our model requires that parameters can be dynamically updated as future similar data emerges. 
To allow room for constantly emerging data, the model parameters are often easily overparameterized \cite{ostapenko2021continual}. 
To regularize the model parameter to maintain a small size and be immune to overfit, we assume that only a small number of neurons should be activated for each distribution. Given the change point variable $\bm{s}_t$, the neural network can dynamically search for a shift-oriented subnetwork. The sparsification rather than using the original network are under two reasons: 1) Lottery Tickets Hypothesis \cite{frankle2018lottery} verifies that a subnetwork can perform as well as a whole network, which is also verified by current findings on model compression; 2) subnetwork can facilitate the learning of evolving network structure, therefor be more robust to distribution shift.

To be specific, for each specific model parameter $\bm{\omega}^{\bm{s}_t}_t$, we decompose it as $\bm{\omega}^{\bm{s}_t}_t = \bm{\omega}_t \odot \bm{m^{\bm{s}_t}}_t$, an element-wise multiplication of a global model parameter $\bm{\omega}_t$ and a shift-driven mask matrix $\bm{m^{\bm{s}_t}}_t$. 
Deriving the exact posterior distribution for the model parameter $\bm{w}$ is intractable due to the nonlinearities of the model. Here, we assume a Gaussian distribution with parameters $\{\bm{\mu}_t^{\bm{s}_t},\mathbf{\Sigma}_t^{\bm{s}_t}\}$ as the posterior distribution of model parameters $\bm{w}$:
 $q_{\bm{\phi}}(\bm{w}_t) = \mathcal{N}(\bm{w}^{\bm{s}_t}|\bm{\mu}^{\bm{s}_t},\bm{\Sigma}^{\bm{s}_t})$.

Following the principle of variational continual learning \cite{nguyen2017variational}, we assume that the prior of $\bm{\omega}_t^{s_k}$ is given by the variational posterior of $\bm{\omega}_{t-1}^{s_k}$, i.e., $\bm{\mu}_t^{\bm{s}_t} = \bm{\mu}_{t-1}^{s_{t-1}}$, $\mathbf{\Sigma}_{t}^{\bm{s}_t} = \mathbf{\Sigma}_{t-1}^{s_{t-1}} $.

Besides, keep in mind that $\bm{\bm{m}_t^{s_{t}}}$ is a learnable model mask parameter that determines which neurons should be activated or set as zeros. To simplify our exposition, we omit script $\bm{s}_{t}$ To model the sparsity property of mask matrix, we define the prior distribution of $p(\bm{m}^{s_{t}})$ using the India buffer process \cite{nguyen2017variational} : $\bm{m} \sim IBP(\alpha)$, where $\alpha$ is the hyperparameters that can control the number of nonzero elements in $\bm{m}$. Specifically, its truncated stick-breaking process for each element in $\bm{m}$ can be denoted as :
\begin{equation} \label{Eq::prior_m}
    v_k \sim Beta(\alpha,1), \pi_k = \prod_{i=1}^{K}v_i, B_{d,k} \sim Bern(\pi_k),
\end{equation}
where $K$ is the truncated level and $\alpha$ controls the value of $K$. To represent the approximated posterior of $m$, we define it as $q(\bm{m}) = q(\bm{m}|v)q(v)$. Corresponding to the Beta-Bernoulli hierarchy of Eq.\ref{Eq::prior_m}, we use the conditional factorized variational posterior family:
\begin{equation}
    q(\bm{m}|v) = \prod_{d-1}^{D}\prod_{k-1}^{K}Bern(m_{d,k}|h_{d,k}),
\end{equation}
where $h_{d,k} = \sigma (\rho_{d,k}+logit(\pi_k))$ and $q(v) = \prod_{k=1}^{K}Beta(v_k|v_{k,1},v_{k,2})$. Consequently, we obtain a set of learnable variational parameters $\{v_{k,1},v_{k,2},\{\mu_{d,k},\Sigma_{d,k},\rho_{d,k}\}\}$. 

\section{Experiment}

\subsection{Experimental Setup}
\subsubsection{Data Description}
We select both synthetic and real-world benchmark dataset to validate the `shift detecting' and `shift adapting' properties of the proposed method. 
Experiments are conducted on 
3 mode system \cite{ansari2021deep} and 
Dancing bees \cite{oh2008learning}. 
Second, we evaluate the ability of the proposed method in adapting to evolving data streams. Experiments are conducted on two tasks, the forecasting task (only the evolving feature $\mathbf{X}_{1:T}$ is given to predict $\mathbf{X}_{t+1}$) and the classification task (learn a mapping from $\mathbf{X}_{t}$ to $\mathbf{Y}_{t}$. Specifically, we use four benchmark forecasting datasets \cite{alexandrov2020gluonts} including traffic, exchange, solar and electricity, where various seasonality patterns, e.g.,  daily, weekly or monthly are shown, and two challenge classification datasets MG\_1C\_2D and optdigits following \cite{souza2015data}.
\subsubsection{Baselines algorithms}
Since there are limited studies working on the same setting with ours, we compared our method with representative continual learning methods, domain adaptation methods via evolving shifted data and other competitive baselines. Specifically, we conduct experiments on both forecasting and classification problems on evolving data. (1) JT (joint training): a naive baseline that train on all the available data ever seen $D_{1,T}$. (2) Adaptive RNN (AdaRNN) \cite{du2021adarnn}: a two-stage method that follow a `segment-adapt' principle for the evolving shifted data. 
(3) Meta-learning via online change point analysis (MOCA) \cite{harrison2020continuous}: an approach which augments a meta-learning algorithm with a differentiable Bayesian change point detection scheme.

\subsection{Experiment Results and Analysis}
\subsubsection{Shift Detection Comparison}
\vspace{-0.15cm}
\begin{table}[h]
\scriptsize
\caption{Comparison of effectiveness on shift detection. Accuracy, NMI, ARI denote the Normalized Mutual Information, the Adjusted Rand Index and segmentation accuracy metrics.} 
\centering
\begin{tabular}{c|c|c|c|c}
\hline
\hline
&&Bouncing Ball & 3 mode system & Dancing bees\\
    \hline
    \multirow{3}{*}{Accuracy} &MOCA
    & $0.93 \pm 0.02$ & $0.94 \pm 0.04 $  & $0.71 \pm 0.03$  \\
     &AdaRNN 
    & $0.93 \pm 0.05$ & $0.90 \pm 0.10$ & $0.68 \pm 0.01$\\ 
     & T-SaS  
    & \textbf{0.96 $\pm$ 0.05}  & \textbf{0.97 $\pm$ 0.00} & \textbf{0.73 $\pm$ 0.14} \\ \hline
    
    \multirow{3}{*}{NMI} &MOCA
    & $0.81 \pm 0.00$  & $0.83 \pm  0.06$ & $0.30 \pm 0.04$ \\
    &AdaRNN 
    & $0.79 \pm 0.01$ & $0.79 \pm 0.13$  & $0.57 \pm 0.01$\\ 
     & T-SaS 
     & \textbf{0.82 $\pm$ 0.011} & \textbf{0.90 $\pm$ 0.02} & \textbf{0.61 $\pm$ 0.01}\\ \hline
     
    \multirow{3}{*}{ARI} &MOCA& $0.85 \pm 0.02$ & $0.89 \pm  0.01$ & $0.38 \pm 0.01$  \\
    &AdaRNN & $0.84 \pm 0.02$ & $0.85 \pm  0.11$ & $0.47 \pm 0.01$\\ 
     & T-SaS  & \textbf{0.86 $\pm$ 0.00} & \textbf{0.94 $\pm$ 0.01} & \textbf{0.54 $\pm$ 0.11} \\ \hline 
     \hline
     \vspace{-0.5cm}
\end{tabular}
\label{Tab:detection}
\end{table}

Experiments in this section are designed to analyze the shift detection behavior of our proposed method.
Experimental results are shown in Table \ref{Tab:detection} and we also represent the visualization in Figure 2 and Figure 3, respectively. Experimental results show that our method achieves significant improvements over the baselines. 
From Figure 2, we observe that AdaRNN apparently struggles  to model long-term sequence data and results in over-segmentation. Additionally, we observe AdaRNN performs a large variance than MOCA and our method. The observation shows its inefficiency to detect the shift point for AdaRNN model, especially for complex sequences. 
It is also obvious that MOCA identifies a fluctuated detection in a three mode system dataset for its inefficiency to model long-term motion patterns.
The observation validates the effectiveness of detecting complex patterns and segmenting the long-term motions for our T-SaS method.
\begin{figure}[h]
\vspace{-0.4cm}
  \centering
\includegraphics[width=\linewidth]{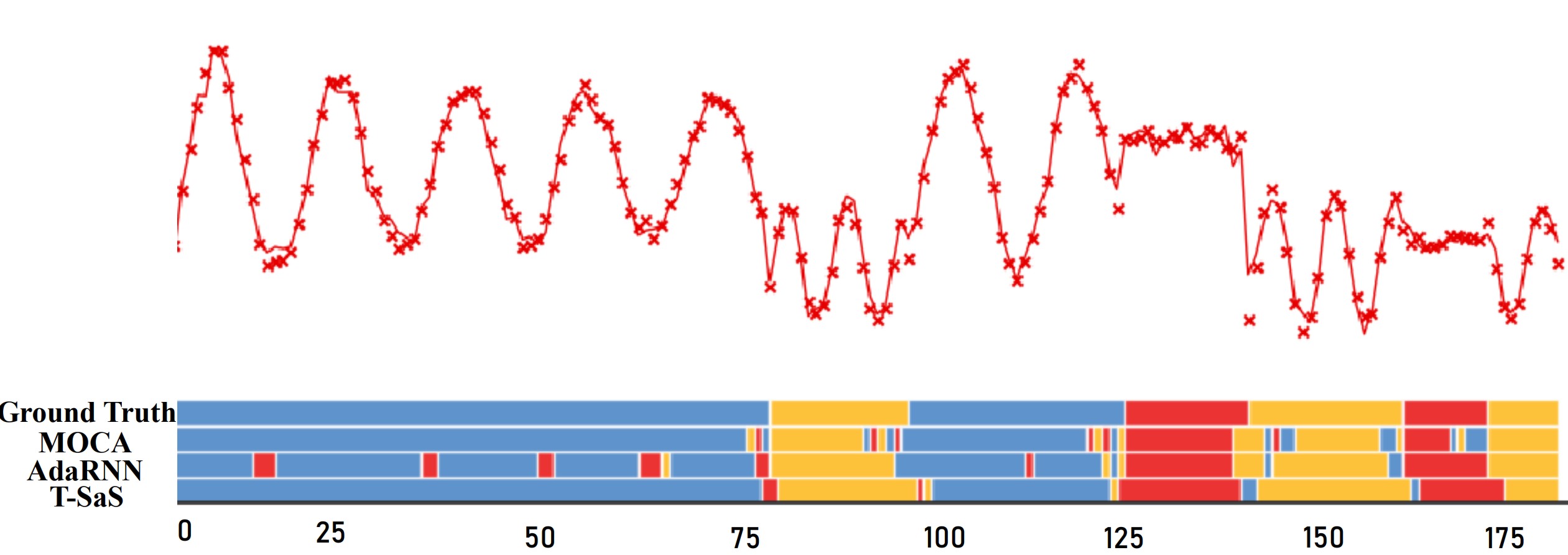}
\caption{Segmentation visualization on the three mode system dataset.}
\label{Fig:seg_3mode}
\vspace{-0.5cm}
\end{figure}
\begin{figure}[ht]
\vspace{-0.5cm}
  \centering
\includegraphics[width=\linewidth]{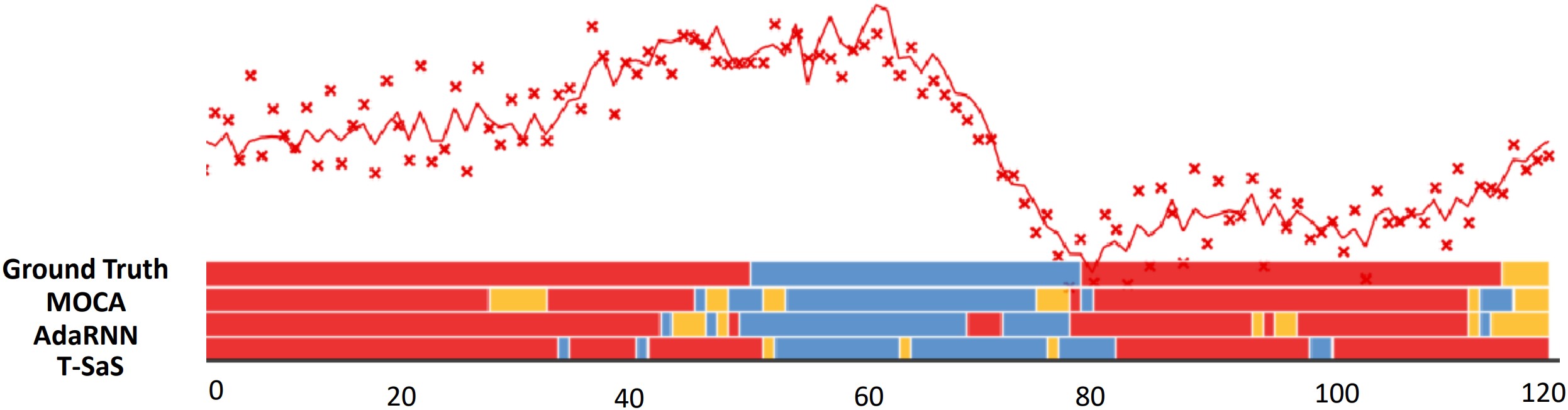}
\caption{Segmentation visualization on the dancing bee dataset.}
\label{Fig:seg_ball}
\end{figure}
\vspace{-0.3cm}
From Figure.3, it is obvious that MOCA fails to identify long segment duration and results in spurious short-term patterns. Differently, our method is able to segment the long-term patterns quite well by taking the advantage of a latent time duration variable. This observation verifies our motivation that a latent time duration variable could benefit the learning of long-term patterns. 
\subsubsection{Drift Adaptation Comparison.}
In this part, experiments are designed to understand if and how the proposed method can adapt to the shifted streaming data.
Experiments on forecasting and classification tasks are shown in Figure 4 and Figure 5, respectively.
\vspace{-0.1cm}
\begin{figure}[h]
  \centering
  \subfloat[Exchange dataset]{
\includegraphics[width=0.23\textwidth,height = 0.13\textheight]{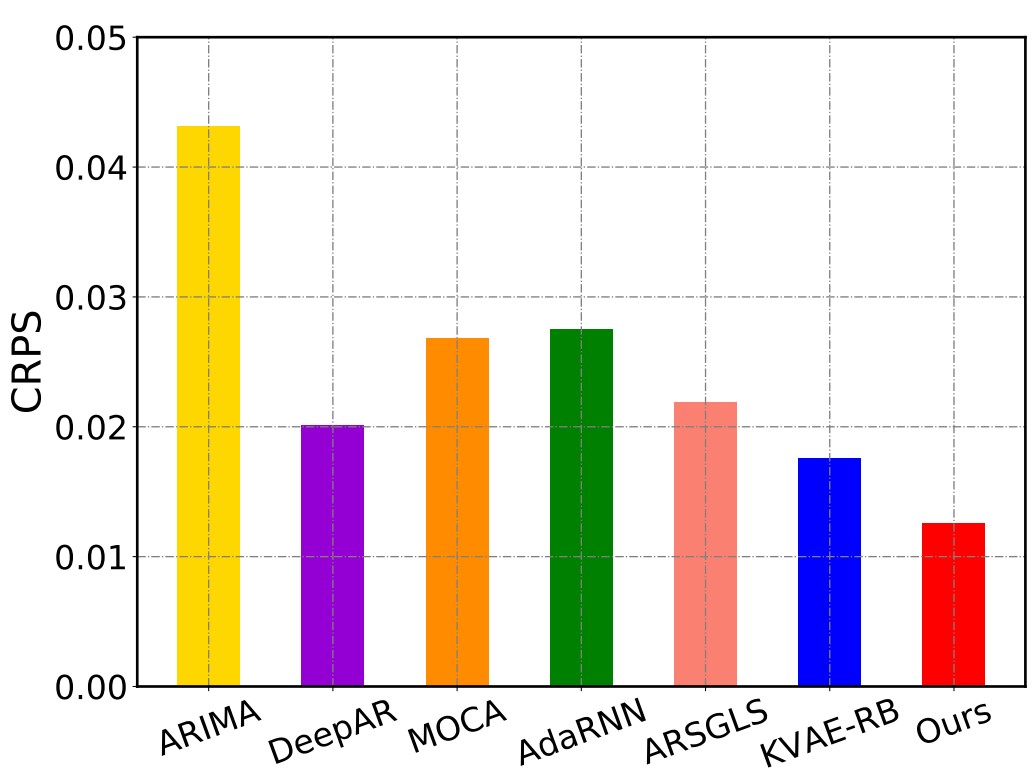}}
  \subfloat[Solar dataset]{
\includegraphics[width=0.23\textwidth,height = 0.13\textheight]{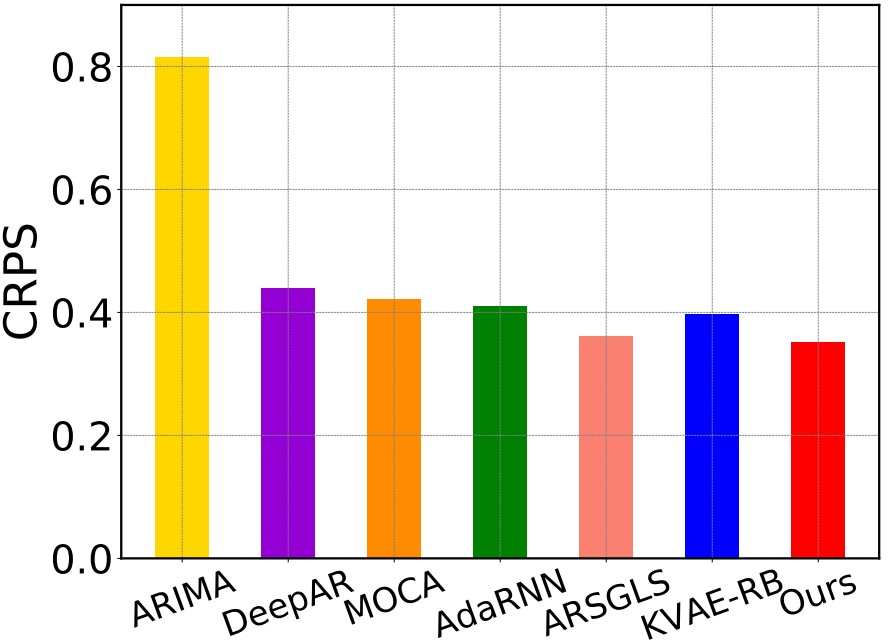}}
\vspace{-0.3cm}
  \subfloat[Electricity dataset]{
\includegraphics[width=0.23\textwidth,height = 0.13\textheight]{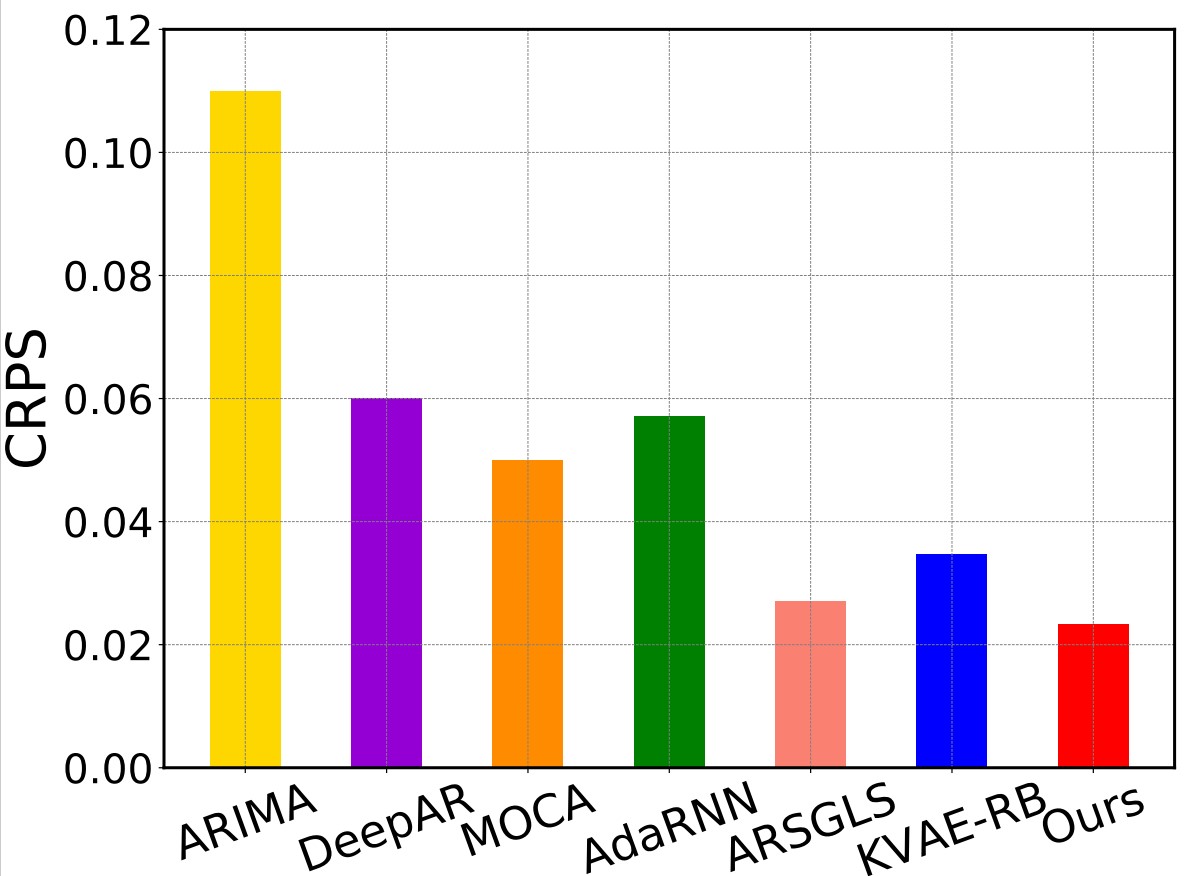}}
  \subfloat[Traffic dataset]{
\includegraphics[width=0.23\textwidth,height = 0.13\textheight]{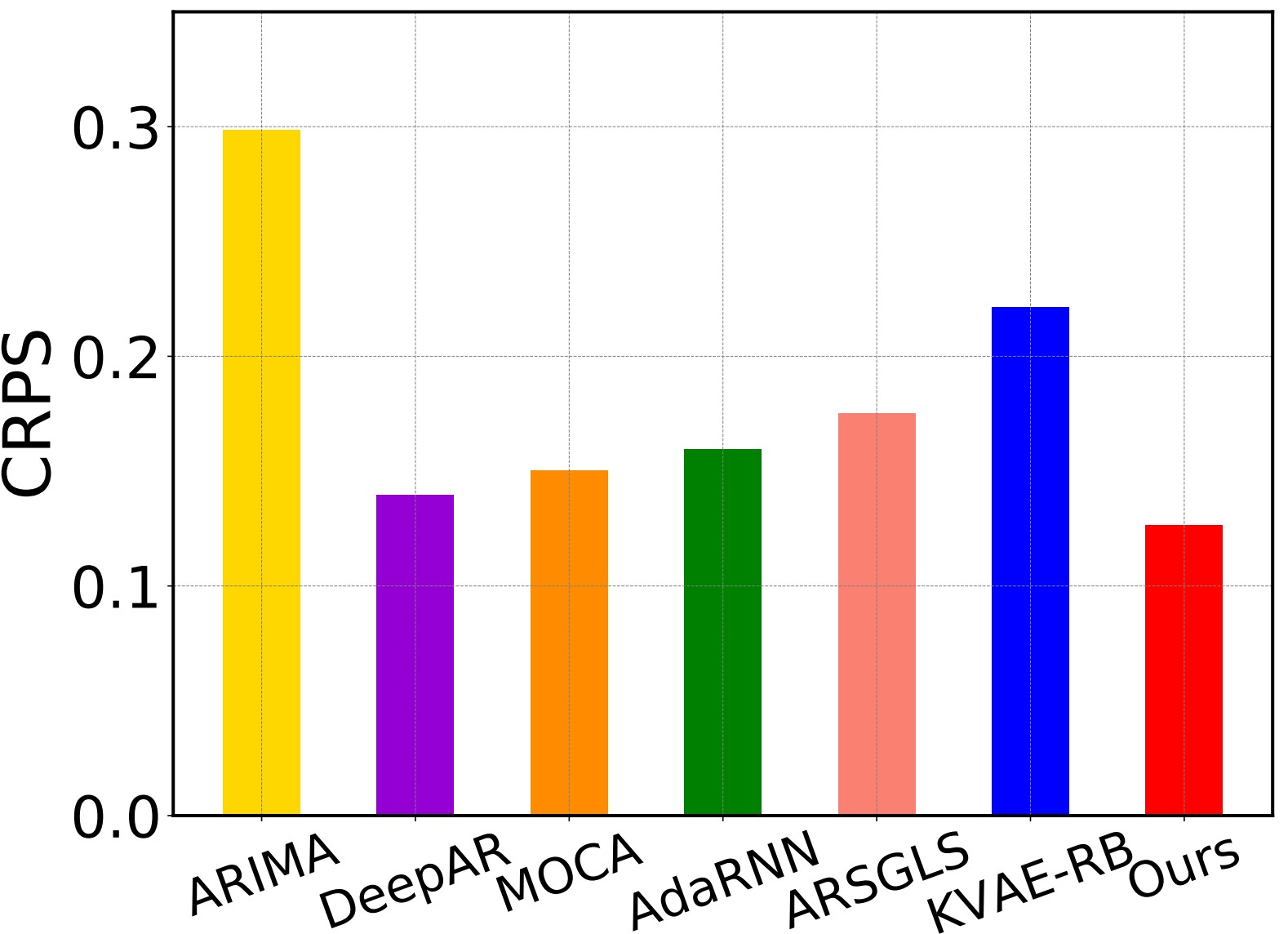}}
  \caption{Forecasting performance across datasets. The lower the value, the better the performance.}
    \label{Fig:forecast}
\vspace{-0.1cm}
\end{figure}
\begin{figure}[h]
\vspace{-0.28cm}
  \centering
  \subfloat[MG\_2C\_2D dataset]{
  \includegraphics[width=0.23\textwidth,height = 0.13\textheight]{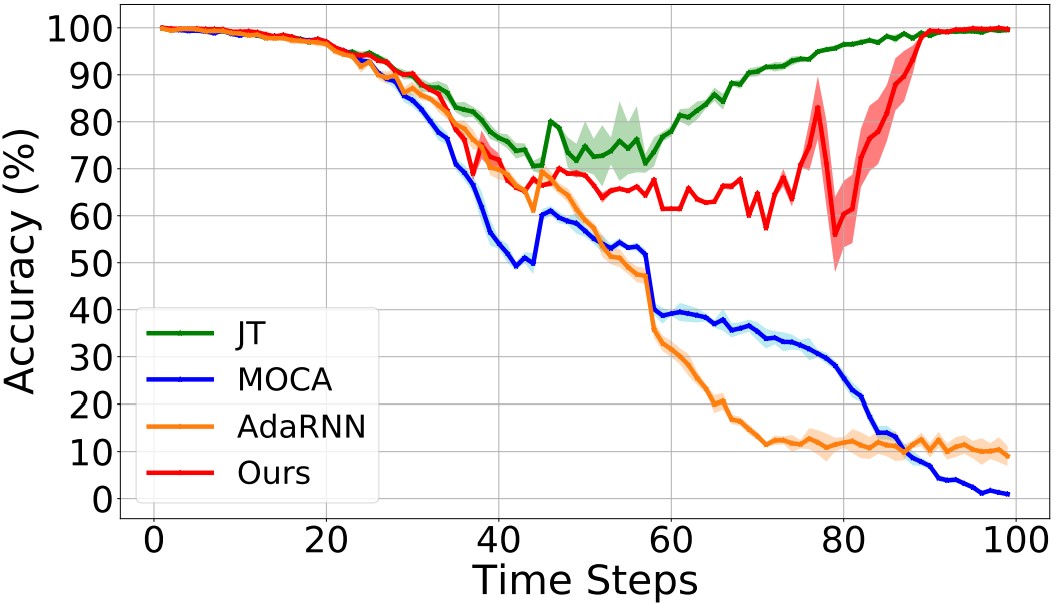}}
  \subfloat[optdigits dataset]{
\includegraphics[width=0.23\textwidth,height = 0.13\textheight]{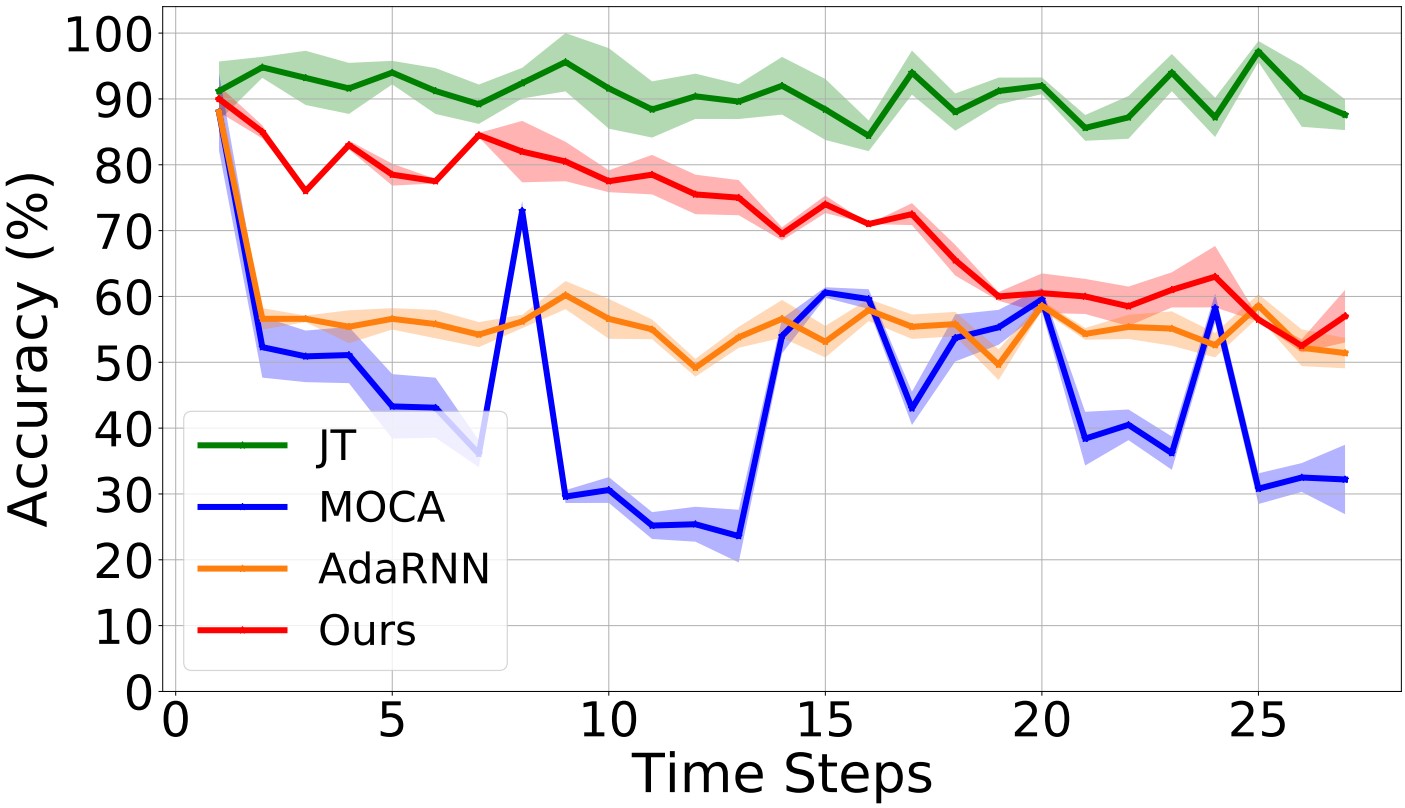}}
  
  \caption{Classification results through timelines.}
    \label{Fig::classification}
\vspace{-0.5cm}
\end{figure}
From Figure 4, We first observe 
our method outperforms all the baselines over all four benchmark datasets. Our method simultaneously infers these distribution shifts and adapts the model to the detected changes with a dynamic model structure. 
Besides, AdaRNN and MOCA show relatively unsatisfactory performance. The main reason is AdaRNN and MOCA adapt to shifted data with global sharing model parameters. Parameter sharing facilitates the transfer of invariance knowledge while penalizing the learning of specialized information. 
Figure \ref{Fig::classification} shows that our method achieves significant improvements over the baselines and is even comparable
with the Joint Training (JT) method (upper bound of the setting). 
Moreover, the margin between our method and baselines, i.e., MOCA and AdaRNN continually increases, indicating our dynamic model architecture can enrich the model expression capacity.

\section{Conclusion}
This paper addresses the problem of modeling low-source streaming data where data shift boundaries are not given.
Additionally, we expand the assumptions regarding distribution shifts to include sudden and irregular patterns.
In addition, we train a specific mask matrix to dynamically route a sparse network from the full network.
We propose an adaptive model for the evolving data by introducing the Bayesian framework with change point variables.
Experimental results on both forecasting and classification tasks demonstrate the effectiveness of the proposed method.


\bibliographystyle{ACM-Reference-Format}
\balance
\bibliography{reference}


\end{document}